\documentclass{icis}

\usepackage{graphicx}
\usepackage{xcolor}
\usepackage{amsmath}
\usepackage{xspace}
\usepackage{longtable}

\title{From EMR Data to Clinical Insight: An LLM-Driven Framework for Automated Pre-Consultation Questionnaire Generation}
\shorttitle{An LLM-driven Automated Pre-Consultation Questionnaire Generation}

\usepackage[hidelinks]{hyperref}
\usepackage{tabularx}  

\begin{document}

\maketitle

\begin{table}[h!]
  \centering
  \begin{tabularx}{\textwidth}{@{}*2{>{\centering\arraybackslash}X}@{}}
    \textbf{Ruiqing Ding}        & \textbf{Qianfang Sun} \\
    Hefei University of Technology & Winning Health Technology Group Co., Ltd.‌  \\
    ruiqingding@hfut.edu.cn & sunqianfang@163.com \\
    \\
    \textbf{Yongkang Leng}        & \textbf{Hui Yin} \\
    Hefei University of Technology & Hefei University of Technology  \\
    2023010136@mail.hfut.edu.cn & yinhui@hfut.edu.cn \\
    \\
    \multicolumn{2}{c}{\textbf{Xiaojian Li}} \\
    \multicolumn{2}{c}{Hefei University of Technology} \\
    \multicolumn{2}{c}{lixj90@hfut.edu.cn} \\
  \end{tabularx}
\end{table}

\begin{abstract}
Pre-consultation is a critical component of effective healthcare delivery. However, generating comprehensive pre-consultation questionnaires from complex, voluminous Electronic Medical Records (EMRs) is a challenging task. Direct Large Language Model (LLM) approaches face difficulties in this task, particularly regarding information completeness, logical order, and disease-level synthesis.
To address this issue, we propose a novel multi-stage LLM-driven framework: 
Stage 1 extracts atomic assertions (key facts with timing) from EMRs;
Stage 2 constructs personal causal networks and synthesizes disease knowledge by clustering representative networks from an EMR corpus;
Stage 3 generates tailored personal and standardized disease-specific questionnaires based on these structured representations.
This framework overcomes limitations of direct methods by building explicit clinical knowledge. 
Evaluated on a real-world EMR dataset and validated by clinical experts, our method demonstrates superior performance in information coverage, diagnostic relevance, understandability, and generation time, highlighting its practical potential to enhance patient information collection.

\end{abstract}

\section{Introduction}
\label{sec:intro}
\begin{figure}[htbp]
	\[
	\begin{array}{|c|}
	\hline \\ [-11pt]
	\includegraphics[scale = 0.35]{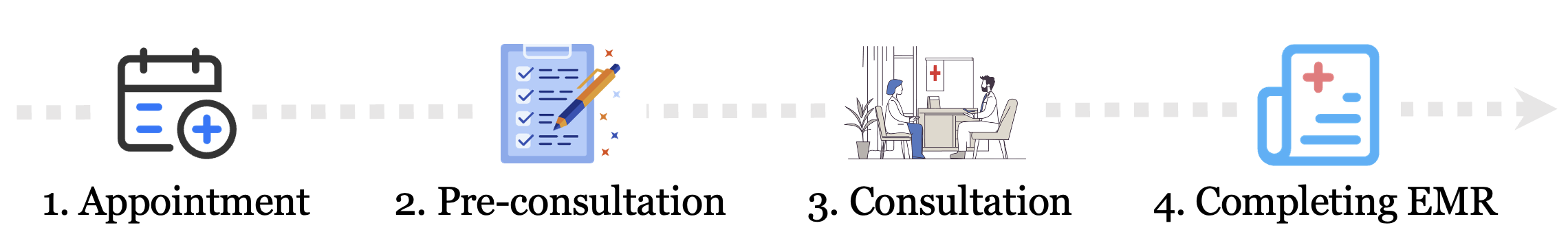} \\ [-4pt]
	\hline
	\addstackgap[7pt]{{\usefont{T1}{ptm}{b}{n}Figure \ref{fig:consultation}.\hspace{0.09cm}Typical Patient Journey including the Pre-consultation Stage.}}\\
	\hline
	\end{array}
	\]
	\captionsetup{labelformat=empty}
	\caption{}\label{fig:consultation} 
\end{figure}

Pre-consultation is recognized as a critical and increasingly vital component of modern healthcare services \autocite{li2024beyond}. 
This process involves gathering essential patient information, such as current symptoms and relevant history, \textit{before} a scheduled clinical visit~\autocite{coallier2017new}. 
Effective pre-consultation streamlines the clinical workflow and helps alleviate the significant administrative burden on physicians. Studies have shown that many physicians spend over 1 hour on electronic health record tasks for every hour of direct clinical face time~\autocite{tai2017electronic}, highlighting the need for solutions that optimize time usage. 
As illustrated in Figure~\ref{fig:consultation}, by providing clinicians with a preliminary understanding of the patient's condition ahead of time, pre-consultation optimizes consultation time and allows for more focused patient-physician interaction. 
This enhanced information flow ultimately contributes to improved diagnostic efficiency, better treatment planning, and a higher quality of patient care. 
Furthermore, in the context of growing healthcare demands and resource disparities worldwide, scalable and effective pre-consultation solutions are becoming increasingly essential \autocite{world2023tracking}.

However, designing comprehensive and effective pre-consultation questionnaires remains an open challenge~\autocite{winston2024medical}.
Conventionally, crafting these questionnaires has relied heavily on the manual effort and clinical expertise of healthcare professionals~\autocite{guyatt1992evidence}.
Although this approach is effective in capturing core symptoms and diagnostic criteria for known conditions, it is inherently time-consuming and labor-intensive. 
Moreover, manually crafted questionnaires struggle to comprehensively cover all essential information. 
This issue is especially pronounced when addressing the complexity and dynamic nature of individual patients' conditions, as well as their diverse medical histories~\autocite{johnson2021precision}.

\vspace{-0.2cm}
\vspace{-1.2cm}

Attempting automation, early work has explored rule-based and decision-tree methods for questionnaire generation~\autocite{ahsan2022machine}.
The fundamental idea behind these approaches is to extract significant information from electronic medical records(EMRs) through keyword matching and utilize pre-defined hierarchical logical paths in decision trees to filter potential symptoms or diagnoses. 
They show certain advantages when symptoms are relatively clear and disease information is fixed, allowing for relatively efficient information classification. 
However, the descriptions from patients are often insufficiently standardized, especially when dealing with vague symptoms and multi-causal relationships, leading to low matching rates. 
Consequently, the logical rigor and coverage of generated questionnaires often fail to meet practical needs.

Beyond rule-based methods, advancements in natural language processing (NLP) have led to intelligent pre-consultation systems utilizing knowledge graphs (KGs)~\autocite{park2021knowledge,li2024unioqa}.
These systems identify entities and relationships within EMRs to facilitate medical inquiries. 
However, constructing comprehensive and up-to-date KGs for the complex medical domain is challenging due to the scarcity of high-quality structured data and the dynamic nature of medical knowledge. 
More crucially, the individualized nature and inherent diversity of real-world patient data found in EMRs make it difficult for KG-based methods to achieve the reliable knowledge mapping required for generating truly tailored pre-consultation questionnaires.

\textbf{Overall, current technological solutions struggle to achieve an ideal balance among the efficiency of questionnaire generation, logical consistency, and the breadth of information coverage. }
Recently, the rapid advancement of large language models (LLMs) has presented new opportunities in NLP tasks, demonstrating remarkable capabilities in understanding, summarizing, and generating complex text, as well as performing various reasoning tasks \autocite{dong2024survey}. 
Leveraging these strengths, LLMs hold significant potential to overcome the limitations of existing methods for processing voluminous, unstructured EMR data and generating comprehensive, context-aware pre-consultation questionnaires. 
In this paper, we propose a novel multi-stage LLM-driven framework designed to automate the generation of comprehensive and clinically relevant pre-consultation questionnaires directly from EMRs. 
Distinct from direct end-to-end LLM applications or simple rule-based systems, our framework employs a structured approach to systematically extract key clinical information, capture complex relationships within EMRs, and synthesize collective knowledge to inform the questionnaire generation process.

We summarize our main contributions as follows:
\begin{itemize}
    \item We propose, to the best of our knowledge, the first multi-stage LLM-driven framework specifically designed for automating medical pre-consultation questionnaire generation from EMRs. This framework moves beyond direct text-to-questionnaire approaches by integrating structured knowledge representation.
    \item We design and elaborate a novel three-stage framework encompassing atomic assertion extraction, causal network construction, and knowledge-informed questionnaire generation. This specific design addresses the critical challenges of handling complex and voluminous EMR data, ensuring comprehensive information capture, logical consistency, and the synthesis of collective clinical knowledge.
    \item We contribute a valuable dataset comprising 3,000 high-quality EMRs collected from a general hospital in Shanghai, China, which is used for evaluation. Through extensive experiments and comparative analysis against baseline methods (manual generation and direct LLM generation), we demonstrate that our proposed framework achieves superior performance across key evaluation metrics including key fact coverage, diagnostic relevance, understandability, and generation time.
\end{itemize}

The remainder of this paper is organized as follows: 
Section 2 reviews related literature; Section 3 describes necessary preliminaries and formulates the research problem; Section 4 details the design of our LLM-driven framework; Section 5 describes the experimental setup and evaluates the framework's performance; and Section 6 discusses the implications and outlines future work.

\section{Related Work}
\label{sec:review}
\textbf{Traditional Methods for EMR Data Mining.}
EMRs serve as the cornerstone for modern clinical decision support systems, encapsulating longitudinal patient histories in structured and unstructured formats. 
Early rule-based approaches (e.g., rule association mining\autocite{zhang2002association}, learning classifier systems\autocite{urbanowicz2009learning}, artificial immune systems \autocite{de2002artificial}) relied on manually constructed decision logic, making them ill-suited for handling the high-dimensional data and unstructured features inherent in complex clinical decision-making. 
KGs demonstrate unique advantages in the field of EMRs through formalized knowledge representation and reasoning mechanisms \autocite{gazzotti2022extending}. 
Research has explored the automatic construction of high-quality KGs directly from EMRs, demonstrating feasibility in this area \autocite{rotmensch2017learning}. 
Building upon this, studies have integrated multi-source heterogeneous drug data (including target pathways and indication associations) with EMRs, validating models for tasks like adverse drug reaction prediction \autocite{bean2017knowledge}. 
An EMR-oriented KG system has also been proposed to integrate fragmented healthcare data and collaboratively support clinical decision-making \autocite{shang2024electronic}. 
Despite these attempts to address EMR challenges using methods like machine learning and KGs \autocite{yaddaden2023machine, likhitha2023developing, yuan2022doctor}, issues such as incomplete medical data, inherent biases, and the unstructured nature of raw data remain major obstacles for traditional AI methods in fully leveraging EMRs.

\textbf{LLMs for Healthcare Applications.}
LLMs have achieved significant progress in various medical domains, including diagnosis \autocite{mcduff2025towards}, patient care \autocite{tripathi2024efficient}, medical literature analysis \autocite{tang2023evaluating}, drug synthesis\autocite{bran2024transformers}, and automated medical record generation \autocite{yang2022gatortron}. 
They provide powerful tools for processing complex medical data and delivering personalized medical recommendations \autocite{thirunavukarasu2023large, clusmann2023future}. 
Task-specific models like BioBERT \autocite{lee2020biobert} and ClinicalBERT \autocite{huang2019clinicalbert} have been developed to address the complexities of clinical language, lexical ambiguities, and unique usage patterns. 
Recent models featuring chain-of-thought (CoT) prompting can further leverage domain expertise and perform complex reasoning \autocite{lievin2024can}. 
Integrating LLM approaches with other methodologies also serves as an effective strategy to enhance clinical capabilities.
For instance, integration strategies include combining LLMs with knowledge graphs to extract symptom-disease relationships and predict diseases \autocite{abdul2024improving}. 
Approaches to enhance natural language understanding and incorporate external knowledge involve leveraging adapters pre-trained on aligning logical representations with natural language \autocite{ni2024knowledge}. 
Generative dialogue systems have also been improved by integrating knowledge graph methods to ensure clinical compliance and human-like medical conversations \autocite{varshney2023knowledge,qiu2024llm}. 
Despite these advancements, directly applying existing LLM approaches to complex, voluminous, unstructured EMRs for automated pre-consultation questionnaire generation presents unique challenges regarding comprehensive information synthesis, logical structuring, and ensuring domain-specific relevance for this particular task.

\textbf{Key Novelty of Our Study.}
Building upon the analysis of existing work, previous methods for pre-consultation questionnaire generation have typically relied on manual design, rule-based systems, or simpler information extraction techniques, often lacking the flexibility and scalability needed to handle diverse and complex patient EMRs. 
Concurrently, despite their promise in other healthcare tasks, directly applying existing LLM approaches to generate questionnaires from unstructured, voluminous EMRs for this specific purpose poses significant challenges concerning comprehensive information completeness, logical order, and disease-specific knowledge synthesis. 
We address this critical gap by proposing a novel multi-stage LLM-driven framework that leverages intermediate structured representations (atomic assertions, causal networks) and synthesizes collective disease knowledge via clustering. 
The key novelty of our framework lies in its departure from direct EMR-to-questionnaire generation, instead introducing and utilizing these intermediate structured representations derived from EMRs, and synthesizing knowledge through network clustering. 
This structured approach, which effectively combines the power of LLMs with explicit knowledge representation, fundamentally differentiates our framework from prior art in both questionnaire generation and general LLM applications in healthcare, offering a more robust and comprehensive solution for automated pre-consultation information gathering.

\section{Problem Formulation}
This section first defines key concepts foundational to our framework, followed by a formal definition of the pre-consultation questionnaire generation task.

\textbf{\MakeUppercase{definition 1}. Atomic Assertion.}
An atomic assertion is defined as the smallest, indivisible semantic unit that precisely represents a single, concrete medical fact or observation extracted from EMRs.
It is structured as an object containing the fact statement and its associated relative time of occurrence. 
For clarity, we use fields such as ``assert" (the fact statement) and ``relative \ time'' (the relative time). 
For example, given the statement, ``the patient has had a persistent headache and fever for 3 days", the corresponding atomic assertions can be represented as: \{\texttt{assert}: ``patient has a persistent headache", \texttt{relative time}: ``3 days ago"\} and \{\texttt{assert}: ``patient has a persistent fever", \texttt{relative time}: ``3 days ago"\}.

\textbf{\MakeUppercase{definition 2}. Personal Causal Network.}
For an individual EMR $M_i$, a personal causal network $G_i=(A_i, E_i)$ is a directed graph where $A_i$ is the set of atomic assertions extracted from $M_i$, and $E_i$ is the set of directed edges representing causal relationships between pairs of assertions in $A_i$. 
An edge $e_{uv} \in E_i$ signifies a causal link from atomic assertion $a_u \in A_i$ to $a_v \in A_i$. This network captures the logical progression and interdependencies of health conditions and events specific to patient $i$.

\textbf{\MakeUppercase{definition 3}. Disease Knowledge Representation.}
For a specific disease $d$, the synthesized knowledge derived from a corpus of EMRs ($\mathcal{M}_d$) is represented as a set of representative causal networks and their corresponding weights $\{[center(C_j), w(C_j)]\}_{j=1}^{m_d}$. 
Each $center(C_j)$ is a representative causal network identified from clustering the personal causal networks of patients with disease $d$, and $w(C_j)$ indicates the prevalence of the clinical pathway represented by $center(C_j)$ within the corpus $\mathcal{M}_d$.
This representation encapsulates the typical causal pathways and their frequencies observed for the specific disease $d$.

\textbf{\MakeUppercase{definition 4}. Pre-consultation Questionnaire.}
A pre-consultation questionnaire $Q$ is a structured collection of questions designed to gather relevant medical information from a patient prior to a clinical consultation. It consists of an ordered series of questions, $Q=\{q_1, q_2, \cdots, q_n\}$, aimed at capturing aspects of the patient's health status, medical history, and related information.

\textbf{\MakeUppercase{problem}. Pre-consultation Questionnaire Generation.}
Given a collection of EMRs $\mathcal{M}=\{M_1, M_2, \\ \cdots, M_N\}$, the problem is to automatically generate pre-consultation questionnaires that are clinically relevant, comprehensive, and tailored. Specifically, the task addresses two main sub-problems:
\begin{enumerate}
    \item \textbf{Personal Questionnaire Generation:} Given an individual previous EMR $M_i$, generate a personal pre-consultation questionnaire $Q_i$ that accurately summarizes the key medical facts and relationships specific to patient $i$ as documented in $M_i$. This is particularly useful for follow-up visits.
    \item \textbf{Disease-Specific Questionnaire Generation:} Given a corpus of EMRs $\mathcal{M}_d \subseteq \mathcal{M}$ for a specific disease $d$, generate a representative pre-consultation questionnaire $\mathcal{Q}_d$ that covers the typical clinical pathways and important factors associated with disease $d$, based on the synthesized knowledge represented by the set of representative networks $\{center(C_j)\}_{j=1}^{m_d}$ and weights $\{w(C_j)\}_{j=1}^{m_d}$. This is primarily intended for first-visit patients with a presumed or confirmed diagnosis of disease $d$.
\end{enumerate}

\section{Methodology}
\label{sec:method}
In this section, we propose an LLM-driven framework to address the pre-consultation questionnaire generation problem. 
We first overview the approach and then illustrate each component of the framework.

\subsection{Overall Framework}
Intuitively, one might consider directly using the full content of EMRs as prompts for LLMs to generate pre-consultation questionnaires.
However, as discussed previously, this straightforward method faces significant limitations.
It struggles to accurately extract all key information and maintain logical order for patients with complex conditions. 
Furthermore, the huge volume of EMR text, especially when attempting to synthesize knowledge for specific diseases from a large corpus, easily exceeds the input length limits of even advanced LLMs.
In this situation, it is infeasible to render a direct end-to-end generation approach for real-world applications.
Based on these considerations, we design a novel multi-stage LLM-driven framework.
As shown in Figure~\ref{fig:framework}, our framework decomposes the complex task into three interconnected stages: 

\vspace{-0.2cm}
\begin{figure}[htbp]
	\[
	\begin{array}{|c|}
	\hline \\ [-11pt]
	\includegraphics[scale = 0.55]{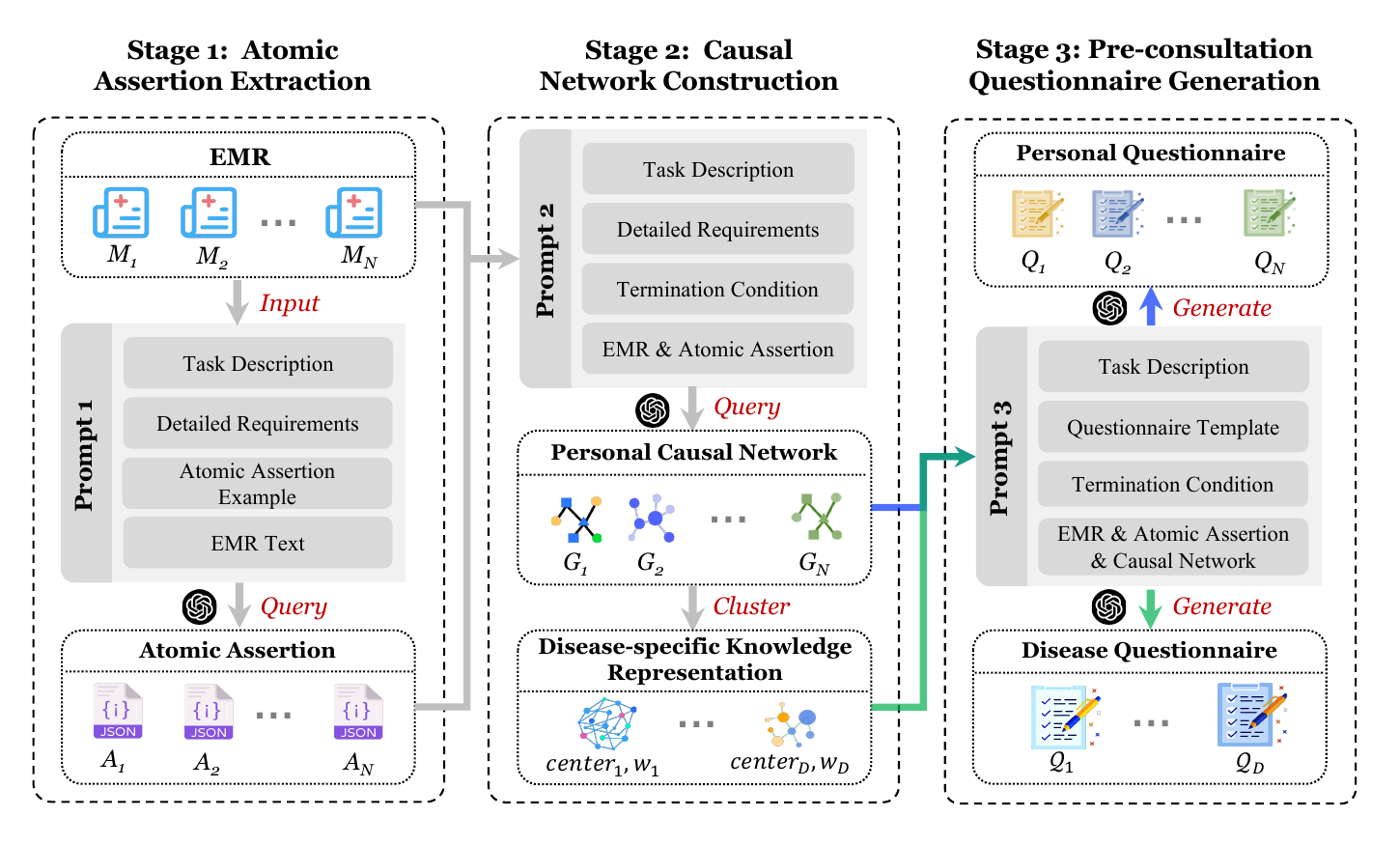} \\ [-4pt]
	\hline
	\addstackgap[7pt]{{\usefont{T1}{ptm}{b}{n}Figure \ref{fig:framework}.\hspace{0.09cm}LLM-Driven Framework for Medical Pre-Consultation Questionnaire Generation}}\\
	\hline
	\end{array}
	\]
	\captionsetup{labelformat=empty}
	\caption{}\label{fig:framework} 
\end{figure}
\vspace{-0.9cm}

\textbf{1. Atomic Assertion Extraction.}
This initial stage processes raw, unstructured EMR text to identify and extract discrete, factual medical statements along with their associated timings, creating a structured representation of atomic assertions (Definition 1). This breaks down complex narratives into manageable, verifiable units.

\textbf{2. Causal Network Construction.}
Building upon the extracted atomic assertions, this stage first constructs a personal causal network for each individual EMR (Definition 2), capturing patient-specific health event dependencies. Subsequently, by analyzing and clustering personal causal networks from a corpus of EMRs for a specific disease, this stage synthesizes a collective disease knowledge representation (Definition 3), reflecting common clinical pathways and their prevalence. 

\textbf{3. Pre-consultation Questionnaire Generation.}
Leveraging the structured representations from the preceding stages, this final stage generates the actual questionnaires. It utilizes the atomic assertions (and potentially personal networks) from Stages 1 and 2 to generate personal questionnaires, while employing the synthesized disease knowledge from Stage 2 to generate disease-specific questionnaires (Definition 4), translating the structured information into clinically relevant and patient-friendly questions using LLMs.

\subsection{Stage 1: Atomic Assertion Extraction}

According to standard medical documentation practices, EMRs are typically composed of specific sections such as chief complaint, history of present illness, systematically documenting the patient's health status, diagnosis and treatment process, and relevant historical information. 
The content within these sections is often presented in unstructured text form, containing complex and rich clinical semantic information. 
Traditional NLP methods for processing EMRs predominantly rely on techniques like entity recognition, 
decomposing text into isolated entities such as symptoms, diagnoses, or medications ~\autocite{park2021knowledge}. 
However, these methods often present significant limitations when dealing with complex contexts and narratives within EMRs, tending to overlook crucial semantic relationships and temporal dependencies. 
This diminishes the coherence and accuracy of extracted information needed for robust clinical understanding.

\vspace{-0.5cm}
\begin{figure}[htbp]
	\[
	\begin{array}{|c|}
	\hline \\ [-11pt]
	\includegraphics[scale = 0.4]{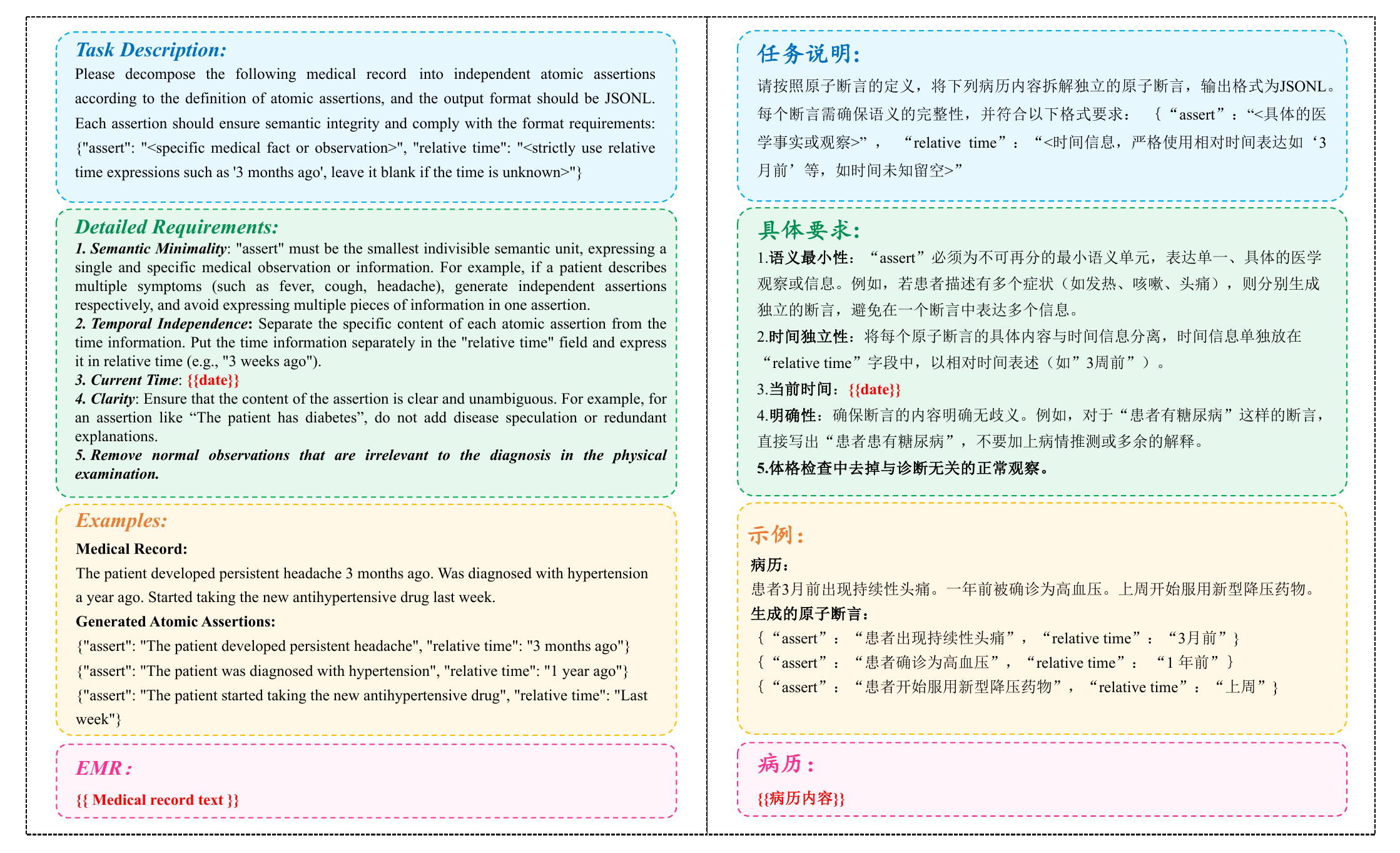} \\ [-4pt]
	\hline
	\addstackgap[7pt]{{\usefont{T1}{ptm}{b}{n}Figure \ref{fig:atomic_assertion}.\hspace{0.09cm}Prompt Template for Atomic Assertion Extraction (left: English, right: Chinese)}}\\
	\hline
	\end{array}
	\]
	\captionsetup{labelformat=empty}
	\caption{}\label{fig:atomic_assertion} 
\end{figure}
\vspace{-0.9cm}

To address the aforementioned issues, we convert the EMR text into atomic assertions with LLM.
As defined previously (Definition 1), an atomic assertion is the smallest, indivisible semantic unit representing a single, concrete medical fact or observation with its associated timing. 
Each unit independently expresses a specific medical fact, observation, or piece of information. 
This method is designed to fully preserve the semantic relationships present in the medical record while providing a refined granularity of information, ensuring completeness and accuracy during the decomposition process.

In this process, the prompt serves as a crucial guiding framework for directing the LLM's extraction behavior. By appropriately describing the requirements for semantic decomposition of the input, explicit separation of temporal information, and focusing the extraction on relevant clinical facts, the prompt guides the LLM to generate a structured set of assertions that conform to medical logic and our defined format. As illustrated in Figure~\ref{fig:atomic_assertion}, a typical prompt primarily consists of a clear task description, detailed output requirements, illustrative examples for few-shot or in-context learning, and the specific medical record content to be processed. This structured prompting approach facilitates task orchestration and helps ensure the accuracy and reliability of the generated atomic assertions across diverse EMR texts.

\subsection{Satge 2: Causal Network Construction}
Medical records meticulously document the patient's symptoms and the diagnostic and treatment process, providing a crucial basis for analyzing the progression of the illness. 
Therefore, constructing a causal network at the individual EMR level can reveal dependencies and influences between medical conditions, assisting clinicians in the prediction and optimization of treatment decisions. 

\begin{figure}[htbp]
	\[
	\begin{array}{|c|}
	\hline \\ [-11pt]
	\includegraphics[scale = 0.4]{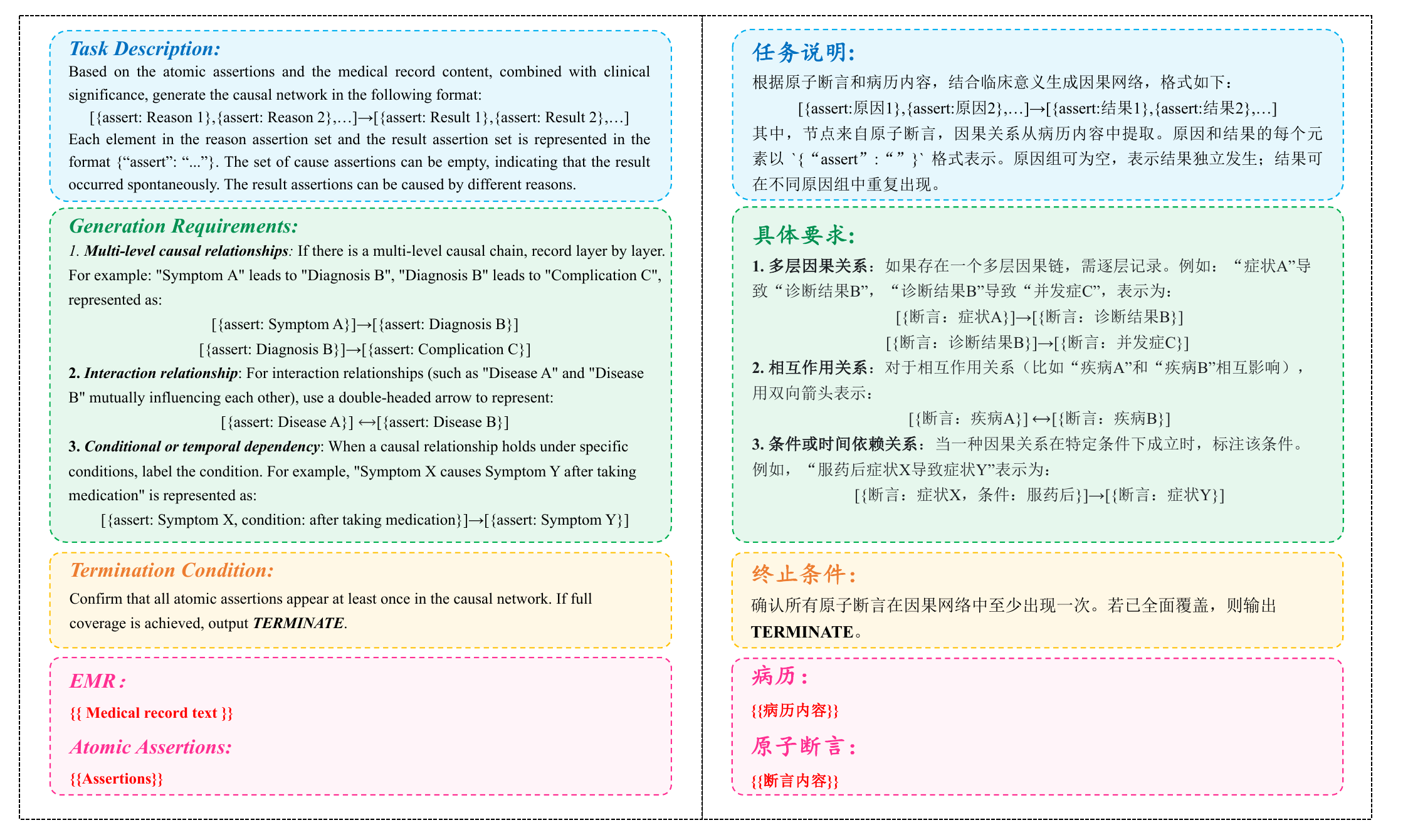} \\ [-4pt]
	\hline
	\addstackgap[7pt]{{\usefont{T1}{ptm}{b}{n}Figure \ref{fig:causal_network}.\hspace{0.09cm}Prompt Template for Causal Network Construction (left: English, right: Chinese)}}\\
	\hline
	\end{array}
	\]
	\captionsetup{labelformat=empty}
	\caption{}\label{fig:causal_network} 
\end{figure}

To achieve this, we design a specific prompt to extract a personal causal network from each individual EMR. 
This process is illustrated in Figure~\ref{fig:causal_network}. 
The prompt takes as input both the original EMR text and the set of atomic assertions previously extracted from it. 
Its purpose is to guide the LLM to identify and extract the relationships, particularly causal and temporal links, that exist between these atomic assertions within the context of the EMR narrative, thereby constructing a network structure. 
Similar to the atomic assertion extraction prompt, the causal network extraction prompt includes a clear task description, detailed generation requirements (e.g., specifying relationship types and output format), criteria for determining the completion of the network, and the input data (EMR text and atomic assertions).

\vspace{-0.5cm}
\vspace{-0.9cm}

For a given set of EMRs $\mathcal{M}=\{M_1, M_2, \cdots, M_N\}$, we first construct a personal causal network $G_i$ for each EMR $M_i$ as described above. 
To derive a generalized representation of the disease's clinical trajectory across different patients, we group the EMRs (and their corresponding personal causal networks) based on the primary diagnosed disease, typically following the ICD-10 code~\autocite{who1992international}. 
For a specific disease $d$, this results in a subset of EMRs $\mathcal{M}_d = \{M_1, \cdots, M_k\}$ and their associated personal causal networks $\{G_1, \cdots, G_k\}$. 
As these personal networks, although capturing unique patient journeys, often contain similar underlying pathways related to symptoms, examinations, and diagnoses for the same disease, despite variations in specific order or expression, we cluster these networks to identify representative disease trajectories. 
This clustering process is essential for synthesizing the collective knowledge from individual cases, filtering out noise and individual peculiarities to reveal the prevalent causal patterns relevant to the disease.

Clustering networks requires a quantitative measure of similarity or distance between them. 
We propose to measure the similarity between two causal networks based on the semantic similarity of their constituent edges, as edges represent the core causal relationships between clinical facts. 
To calculate the similarity among the networks, we first utilize BCEmbedding~\autocite{youdao_bcembedding_2023} to convert the semantic content of each edge into a vector embedding. 
Specifically, for an edge $e_{ij}$ connecting two nodes represented by atomic assertions $a_i$ and $a_j$, the edge embedding $v(e_{ij})$ is obtained by concatenating the embeddings of the two nodes:
\begin{equation}
    v(e_{ij}) = \text{Concat}(\text{BCEmbedding}(a_i), \text{BCEmbedding}(a_j))
\end{equation}
This concatenation strategy is chosen because the meaning of a causal relationship is inherently linked to the semantic content of both the cause ($a_i$) and the effect ($a_j$).
Given two networks $G_k$ and $G_l$, we define the similarity between them, $\text{sim}(G_k, G_l)$, as the average of the cosine similarities between all possible pairs of edges, one from $G_k$ ($e_{ij} \in E_k$) and one from $G_l$ ($e_{mn}\in E_l$):
\begin{equation}
    \text{sim}(G_k, G_l) = \frac{1}{|E_k| \cdot |E_l|} \sum_{e_{ij} \in E_k} \sum_{e_{mn}\in E_l} \text{sim}(e_{ij}, e_{mn})
\end{equation}
where $\text{sim}(e_{ij}, e_{mn})=\frac{v(e_{ij}) \cdot v(e_{mn})}{|v(e_{ij})| \cdot |v(e_{mn})|}$ is the cosine similarity between the embeddings $v(e_{ij})$ and $v(e_{mn})$. 
Calculating the average similarity across all edge pairs allows $\text{sim}(G_k, G_l)$ to reflect the overall semantic similarity of the causal structures represented by the two networks, providing a robust measure for clustering.

With a similarity measure defined, we apply the hierarchical clustering algorithm to group the networks $\{G_1, \cdots, G_k\}$ belonging to the same disease into clusters $C = \{C_1, \cdots, C_m\}$. 
Hierarchical clustering is chosen for its ability to reveal the multi-scale structure of relationships between networks~\autocite{murtagh2014ward}. 
We use average linkage to determine the distance between two clusters $C_i$ and $C_j$:
\begin{equation}
    d(C_i, C_j) = \frac{1}{|C_i| \cdot |C_j|}\sum_{G_k \in C_i} \sum_{G_l \in C_j} (1-\text{sim}(G_k, G_l))
\end{equation}
where $(1-\text{sim}(G_k, G_l))$ converts the similarity measure into a distance measure. 
We then obtain the final clusters by cutting the dendrogram at a fixed cut-off value, where each resulting cluster $C_i$ contains a set of personal causal networks that are structurally and semantically similar, representing a distinct typical pathway.

From each cluster $C_i$, which represents a typical disease pathway observed in the data, we select a central network $center(C_i)$ to serve as its representative structure. 
The central network is chosen as the network within the cluster that minimizes the sum of distances to all other networks in that cluster:
\begin{equation}
    center(C_i) = \arg \min_{G_k \in C_i} S(G_k)
\end{equation}
where $S(G_k)$ is the sum distance of $G_k$ to the other networks in the same cluster, calculated as:
\begin{equation}
    S(G_k) = \sum_{G_j \in C_i} (1 - \text{sim}(G_k, G_j))
\end{equation}
This method ensures that the chosen $center_{C_i}$ is the most representative network structure within its cluster. 
Besides selecting a representative, we also calculate the weight of each cluster $C_i$ to indicate its prevalence or significance among the networks for that disease. The weight $w_{C_i}$ is defined as the proportion of personal networks belonging to cluster $C_i$ relative to the total number of networks for the disease $d$:
\begin{equation}
    w(C_i) = \frac{|C_i|}{\sum_{j=1}^m |C_j|}
\end{equation}
where $m$ is the total number of clusters identified for disease $d$. This weight reflects how frequently the clinical pathway represented by the cluster $C_i$ is observed in the dataset.

Finally, for each disease, this comprehensive process yields a set of representative causal networks and their corresponding weights $ \{[center(C_i), w(C_i)]\}_{i=1}^{m}$. 
This set collectively describes the typical causal pathways and their prevalence for the specific disease, providing a consolidated, data-driven knowledge base for the subsequent questionnaire generation stage.

\subsection{Stage 3: Pre-consultation Questionnaire Generation}
The final stage of our framework is to generate the pre-consultation questionnaires, where the structured medical information and synthesized knowledge are transformed into user-friendly questionnaires ready for clinical application. 
This stage is designed to produce two distinct types of questionnaires to meet different clinical needs: personal questionnaires and disease-specific questionnaires.

\textit{\textbf{A personal questionnaire}} is generated for an individual patient using their single EMR, leveraging extracted atomic assertions and the constructed personal causal network ($G_i$). 
An LLM is prompted with this processed individual data to translate the patient's specific medical facts and relationships into clear questions. 
The prompt ensures coverage of all assertions, predominantly using multiple-choice with options for free-text input, and organizes questions logically based on the personal network structure.
\textit{\textbf{A disease-specific questionnaire}} is a standardized tool generated for patients with a particular disease. 
It is based on the collective knowledge synthesized from an EMR corpus in Stage 2, specifically the set of representative disease causal networks ($\{center_{C_j}\}_{j=1}^{m}$) and their associated weights ($\{w_{C_j}\}_{j=1}^{m}$). 
A dedicated prompt guides an LLM to synthesize these typical pathways and prevalent factors into a comprehensive, general questionnaire, prioritizing content based on network weights and maintaining a logical flow.

\begin{figure}[htbp]
	\[
	\begin{array}{|c|}
	\hline \\ [-11pt]
	\includegraphics[scale = 0.4]{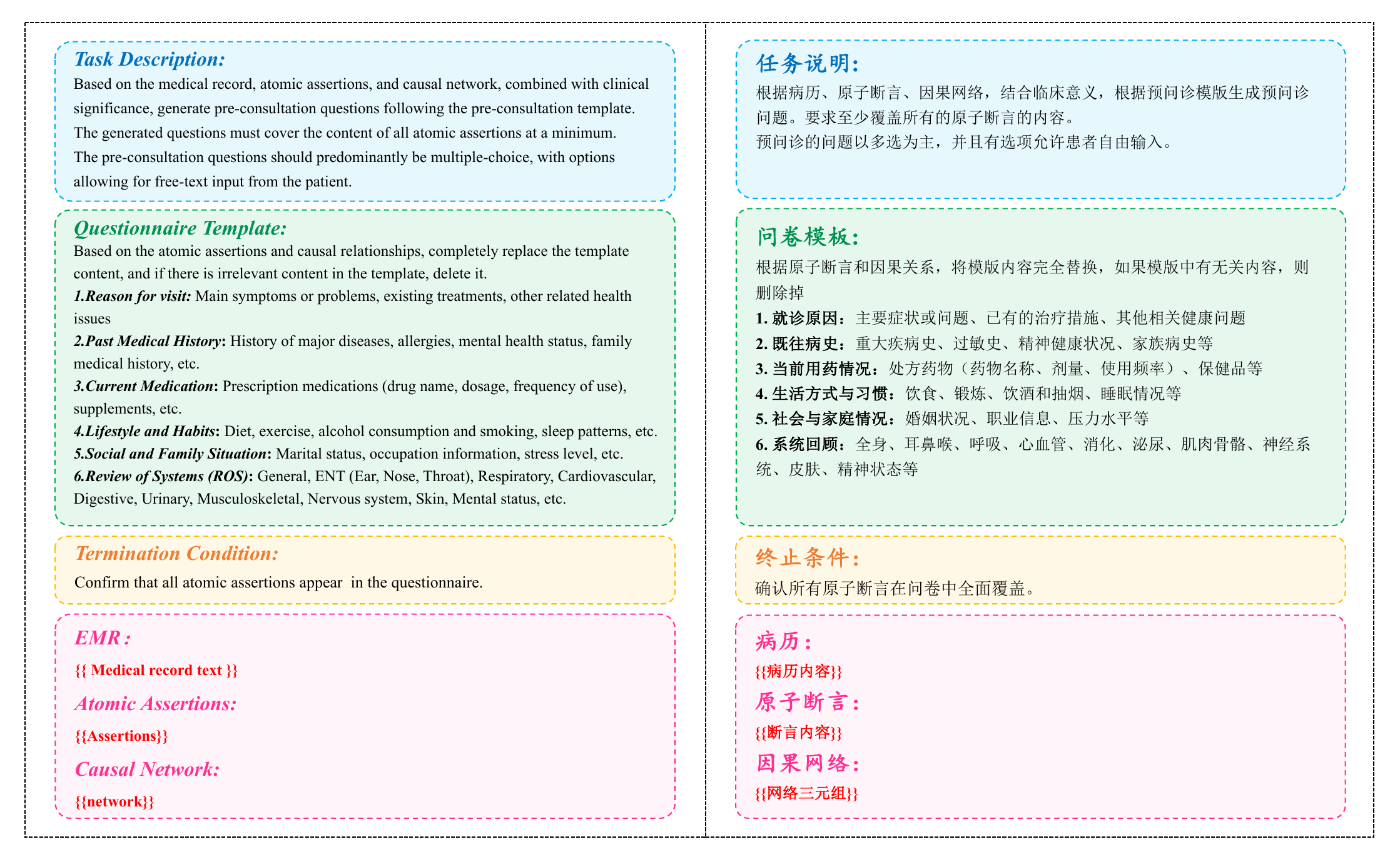} \\ [-4pt]
	\hline
	\addstackgap[7pt]{{\usefont{T1}{ptm}{b}{n}Figure \ref{fig:questionnaire_generation}.\hspace{0.09cm}Prompt Template for Questionnaire Generation (left: English, right: Chinese)}}\\
	\hline
	\end{array}
	\]
	\captionsetup{labelformat=empty}
	\caption{}\label{fig:questionnaire_generation} 
\end{figure}

The general structure of the prompts used for generating both personal and disease-specific questionnaires is similar, typically including a task description, requirements for output format and content, and specifications for input data, as illustrated in Figure~\ref{fig:questionnaire_generation}.

\vspace{-0.5cm}
\vspace{-0.9cm}

In both personal and disease-specific questionnaire generation processes, the LLM plays the crucial role of bridging the gap between structured medical knowledge representations (assertions, networks, representative pathways) and natural language questions, ultimately producing the final output used in clinical practice.

\section{Experiments}
\label{sec:results}

In this section, we evaluate our method to ascertain its effectiveness in pre-consultation questionnaire generation.
Our investigation is guided by three pivotal research questions:

\textbf{RQ1}: Can the proposed framework effectively demonstrate the full process of converting EMR data through its multi-stage pipeline into a pre-consultation questionnaire?

\textbf{RQ2}: For personal pre-consultation questionnaire generation, does our framework achieve superior performance compared to directly prompting an LLM?

\textbf{RQ3}: For disease-specific pre-consultation questionnaire generation, does our framework achieve superior performance compared to manual generation by clinical experts?

\subsection{Experiment Settings}
\label{sec:design}

\textbf{Datasets.}
The experimental data were collected from 3,000 de-identified EMRs obtained from a general hospital in Shanghai, China, spanning the period from January 2023 to June 2024. This dataset encompasses records from various departments and covers diverse disease types, with a significant portion containing information on multiple concurrent diseases per patient, providing a rich basis for causal network analysis.

All collected EMRs underwent rigorous quality control and contain comprehensive details on symptoms, diagnoses, and treatment plans. 
Prior to experiments, the raw text was preprocessed to clean noise and irrelevant information, aiming to facilitate subsequent analysis.
Statistical information about the dataset is presented in Table~\ref{tab:data_sta}.

\textbf{Evaluation Metrics.}
To evaluate the effectiveness of our proposed method, we employ four key metrics: key fact coverage ($C$), diagnostic relevance ($R$), understandability ($U$), and generation time ($T$). These metrics are defined as follows:

\vspace{-0.9cm}
\setlength{\LTleft}{-4pt}
\bgroup
\def\arraystretch{1.3}
\footnotesize
\vspace{0.8cm}
\begin{table}[ht]
	\centering
	\begin{tabular}{|c|c|c|}
        \hline
	      \textbf{Department}  & \textbf{EMR Numbers} & \textbf{Common Diseases} \\ \hline
		Dermatology 	   & 750   & Eczema, psoriasis, dermatitis           \\ \hline
		Pulmonology   	   & 750   & Bronchitis, pneumonia, tuberculosis     \\ \hline
		Internal Medicine  & 750   & Hypertension, diabetes, heart disease   \\ \hline 
            Others             & 750   & Gastric diseases, kidney diseases, etc. \\ \hline
		\multicolumn{3}{|c|}{\normalsize{{\usefont{T1}{ptm}{b}{n}Table \ref{tab:data_sta}. \hspace{0.09cm} Basic Information of the Collected EMRs.}}} \rule{0pt}{3ex} \\ [4pt] \hline
	\end{tabular}
	\captionsetup{labelformat=empty}
	\caption{}
	\label{tab:data_sta}
\end{table}
\egroup
\vspace{-0.9cm}

\begin{itemize}
\item \textbf{Key Fact Coverage} ($C$).
This metric quantifies the proportion of relevant clinical facts from the EMRs that are captured in the generated questionnaire, indicating the comprehensiveness of the questionnaire.
A higher coverage rate signifies that the questionnaire incorporates more of the essential information available in the source.
We define two variations of this metric:
\begin{itemize}
    \item \textbf{Personal Key Fact Coverage} ($C_{\text{personal}}$): Calculated for questionnaires generated from individual EMRs, measuring the proportion of key facts present in the single patient's EMR that appear in the generated personal questionnaire.
    \item \textbf{Disease Key Fact Coverage} ($C_{\text{disease}}$): Calculated for the common questionnaire generated for a specific disease from a corpus of EMRs, measuring the proportion of key facts extracted from this corpus that are included in the specific disease questionnaire.
\end{itemize}

\item \textbf{Relevance to Diagnosis} ($R$).
This metric quantifies how well the generated questionnaire's content aligns with the needs for diagnosing the specific disease. It is assessed by clinical experts based on whether the questions conform to established clinical norms and common sense, effectively aid in determining the disease diagnosis, and exhibit overall clinical practicality. Scores are assigned on a scale from 0 to 10, with 10 representing the highest relevance.

\item \textbf{Understandability} ($U$).
This metric evaluates the clarity with which the questionnaire's content and the rationale underpinning its design can be comprehended by its intended users. Clinical experts assess whether the questionnaire effectively conveys the purpose, background, and generation basis of each question to doctors, patients, and other relevant personnel. This metric is scored on a scale from 0 to 10, where 10 represents the highest level of understandability.

\item \textbf{Generation Time} ($T$).
This metric quantifies the time required to generate a questionnaire using a particular method, serving as a measure of computational efficiency. It is used to compare different approaches, with a shorter generation time indicating higher efficiency and therefore being preferred.
\end{itemize}

For the Relevance to Diagnosis ($R$) and Understandability ($U$) metrics, evaluation was performed by a panel of 5 clinical experts. 

\vspace{-0.2cm}
\begin{figure}[htbp]
	\[
	\begin{array}{|c|}
	\hline \\ [-11pt]
	\includegraphics[scale = 0.37]{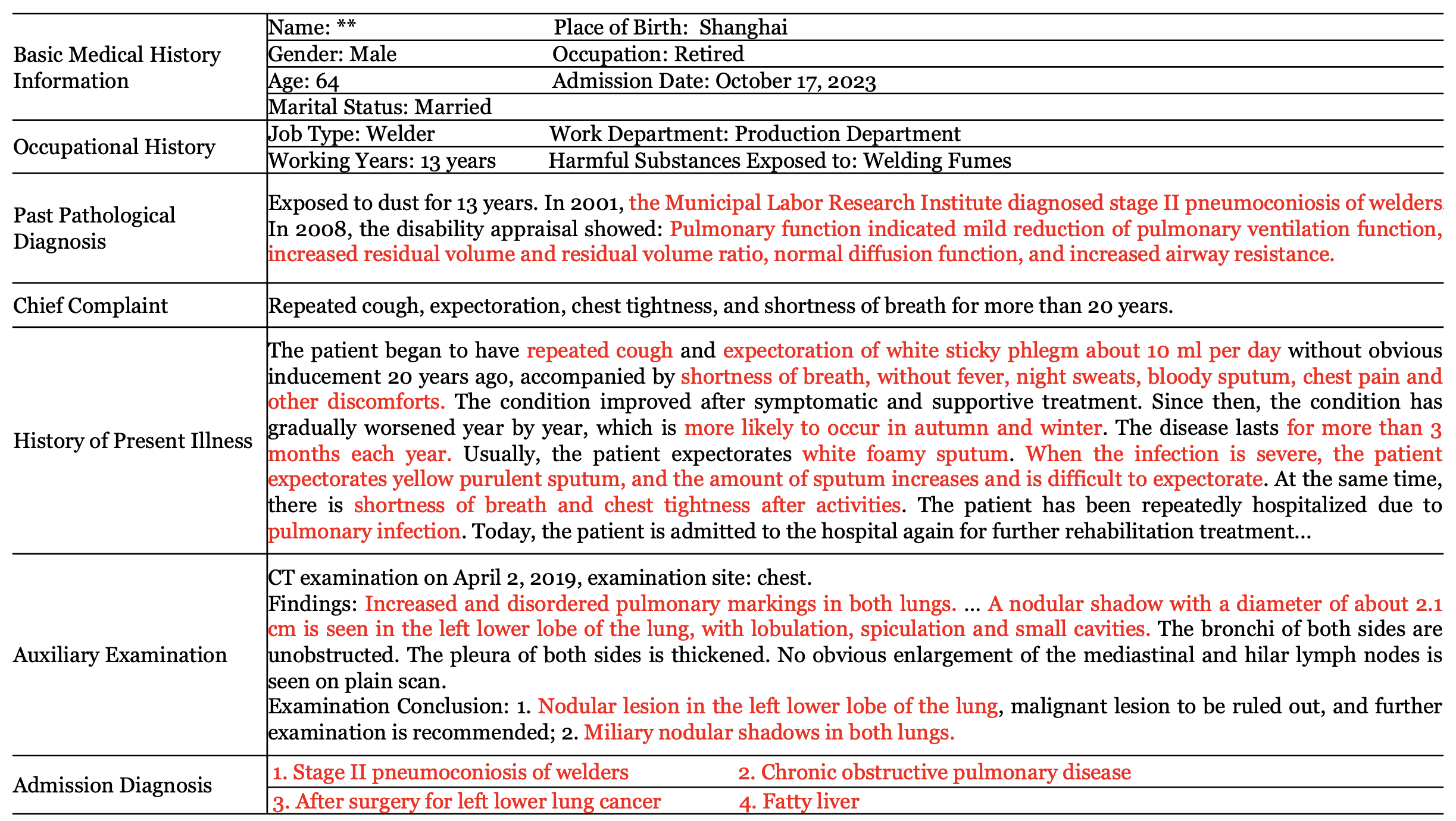} \\ [-4pt]
	\hline
	\addstackgap[7pt]{{\usefont{T1}{ptm}{b}{n}Figure \ref{fig:res_emr}.\hspace{0.09cm}A Portion of the Patient's EMR (Important medical facts are in red.)}} \\
	\hline
	\end{array}
	\]
	\captionsetup{labelformat=empty}
	\caption{}\label{fig:res_emr} 
\end{figure}
\vspace{-0.9cm}

\textbf{Compared Methods.}
To benchmark the performance and effectiveness of our proposed framework, we compare it against 2 representative baseline methods commonly used for generating pre-consultation questionnaires: 
\begin{itemize}
    \item \textbf{Manual generation by clinical experts }(\textit{Manual}).
    For this baseline, we invited 5 clinical experts with questionnaire writing experience to manually craft questionnaires based on their medical knowledge and clinical practice. 
    While this approach often ensures high clinical quality and relevance, it is inherently time-consuming, resource-intensive, and challenging to scale efficiently for a wide range of diseases or patient variations.
    \item \textbf{Direct LLM generation from EMRs }(\textit{LLM}).
    This baseline method applies an LLM directly to the raw text of EMRs to generate questionnaires. 
    While offering the potential for automation, this approach often struggles with processing complex and lengthy EMR texts due to context window limitations and may fail to reliably extract all key information or maintain correct chronological order, particularly when synthesizing information across multiple records for a specific disease.
\end{itemize}

\textbf{Implementation Details.}
All experiments are conducted on machines running the Ubuntu 20.04 operating system, utilizing CUDA version 12.1 for GPU acceleration, and equipped with NVIDIA GeForce RTX 4090 GPUs. 
The software environment included Python 3.9, PyTorch 2.5.1, and scikit-learn 1.6.0. 
As the core LLM component, we select GPT-4o~\autocite{OpenAI_GPT4o} for its advanced capabilities in our experiments. 

\subsection{Experiment Results}
Here, we present the experimental results evaluating the effectiveness and performance of our proposed framework for pre-consultation questionnaire generation, addressing the research questions outlined above.

\textbf{Demonstration of the Framework Process(RQ1)}

We selected a sample EMR from our dataset that clearly illustrates the extraction of key facts, the construction of relationships, and the subsequent generation of relevant questions. 
Figure~\ref{fig:res_emr} shows a portion of the original unstructured EMR text used as the input for this example walkthrough.
The original EMR text was in Chinese and has been translated into English for this demonstration.
It contains various medical facts embedded within clinical narratives, which are marked in red.

From this input text, Stage 1 extracts a set of discrete atomic assertions, representing the key medical facts and their associated timings. 
A selection of these extracted atomic assertions from the example EMR is presented below in JSONL format. 
These assertions provide a structured, granular representation of the patient's condition narrative.

\vspace{0.2cm}
\hrule
\vspace{-0.4cm}
\begin{verbatim}
{"assert": "Patient has Welder's Pneumoconiosis Stage II", "relative time": ""},
{"assert": "Patient has Chronic Obstructive Pulmonary Disease", "relative time": ""},
...
{"assert": "Patient developed recurrent cough", "relative time": "Over 20 years ago"},
{"assert": "Patient has no fever", "relative time": ""},
{"assert": "Patient's chest CT shows a nodule in the left lower lobe", "relative time": 
  "4.5 years ago"}
\end{verbatim}
\vspace{-0.2cm}
\hrule

\begin{figure}[htbp]
	\[
	\begin{array}{|c|}
	\hline \\ [-11pt]
	\includegraphics[scale = 0.28]{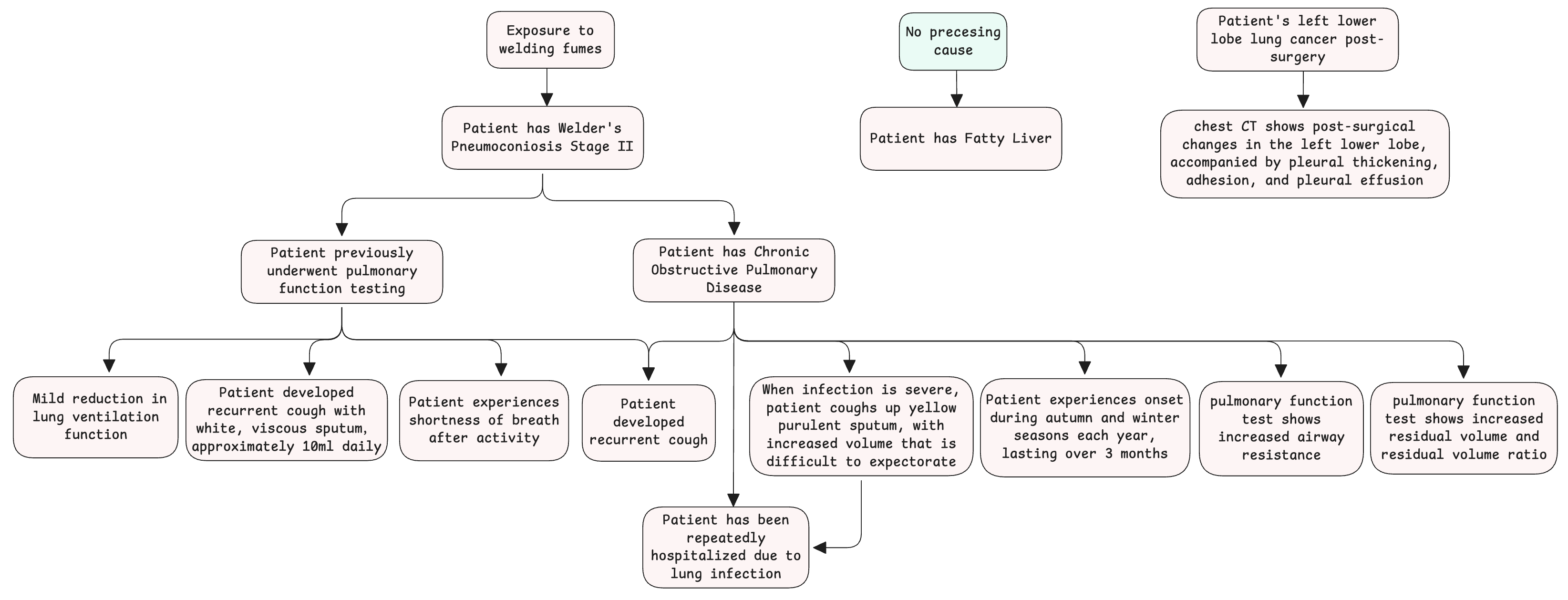} \\ [-4pt]
	\hline
	\addstackgap[7pt]{{\usefont{T1}{ptm}{b}{n}Figure \ref{fig:personal_network}.\hspace{0.09cm}Visualization of a Portion of the Personal Causal Network}} \\
	\hline
	\end{array}
	\]
	\captionsetup{labelformat=empty}
	\caption{}\label{fig:personal_network} 
\end{figure}

These extracted atomic assertions, analyzed in the context of the original EMR text, are then used in Stage 2 to construct the personal causal network for this patient. 
This network is to explicitly model the relationships (e.g., causal, temporal) between the atomic assertions, capturing the progression and interdependencies of the patient's health conditions as described in the EMR. Figure~\ref{fig:personal_network} provides a visualization of a portion of the personal causal network constructed for this example EMR, showing how atomic assertions are connected.

\vspace{-0.2cm}
\vspace{-0.9cm}

\begin{figure}[htbp]
	\[
	\begin{array}{|c|}
	\hline \\ [-11pt]
	\includegraphics[scale = 0.35]{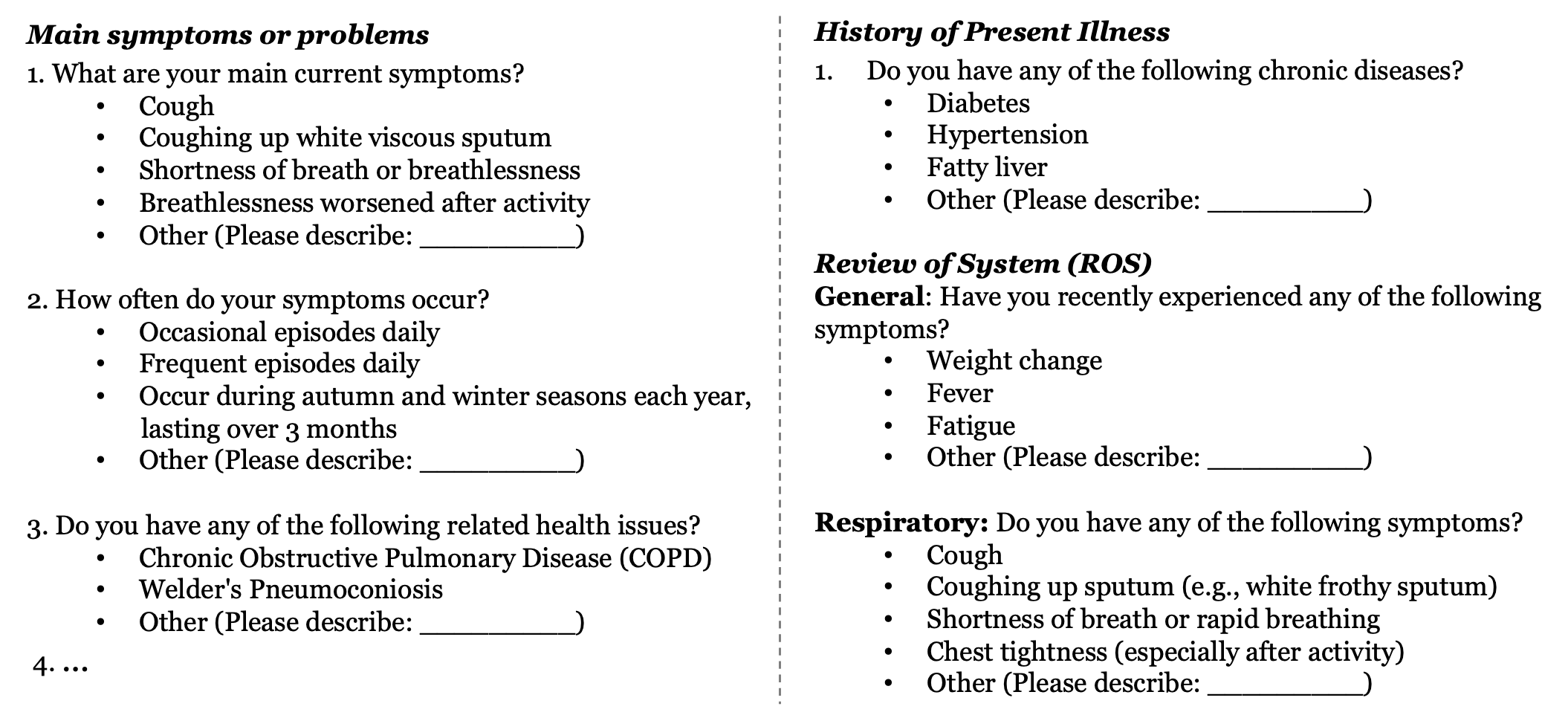} \\ [-4pt]
	\hline
	\addstackgap[7pt]{{\usefont{T1}{ptm}{b}{n}Figure \ref{fig:personal_questionnaire}.\hspace{0.09cm}A Portion of the Generated Personal Questionnaire}} \\
	\hline
	\end{array}
	\]
	\captionsetup{labelformat=empty}
	\caption{}\label{fig:personal_questionnaire}
\end{figure}

Finally, in Stage 3, the personal causal network, along with the extracted atomic assertions, serves as input for generating a tailored pre-consultation questionnaire for this patient. 
This stage translates the structured knowledge and relationships derived in the preceding stages into a series of patient-friendly questions designed to verify and elaborate on the EMR's content. 
Key sections of the generated personal questionnaire, directly derived from the preceding stages, are shown in Figure~\ref{fig:personal_questionnaire}.

As this example walkthrough demonstrates, the proposed framework effectively executes the full pipeline from raw EMR data through intermediate structured representations to the final generation of a pre-consultation questionnaire.

\textbf{Performance Evaluation for Personal Questionnaires (RQ2)}

To illustrate the practical process and qualitatively demonstrate the performance differences, we first present a detailed example using the EMR shown in Figure~\ref{fig:res_emr}. 
This sample EMR contains a total of 38 key clinical facts. 
Our proposed framework generates a personal questionnaire that successfully covers 32 of these key facts. 
In sharp contrast, a direct LLM approach, using the same EMR text as input, covered only 16 facts. 
This substantial disparity in coverage from a single complex case qualitatively demonstrates the inherent challenges of directly applying LLMs to extract and structure information comprehensively from unstructured clinical narratives, leading to incomplete coverage of important medical facts.
Our method, by effectively leveraging extracted atomic assertions and constructed causal networks, shows a clear advantage in comprehensively capturing and reflecting the core elements contained within individual EMRs.

\vspace{-0.2cm}
\vspace{-0.9cm}

\begin{figure}[htbp]
	\[
	\begin{array}{|c|}
	\hline \\ [-11pt]
	\includegraphics[scale = 0.3]{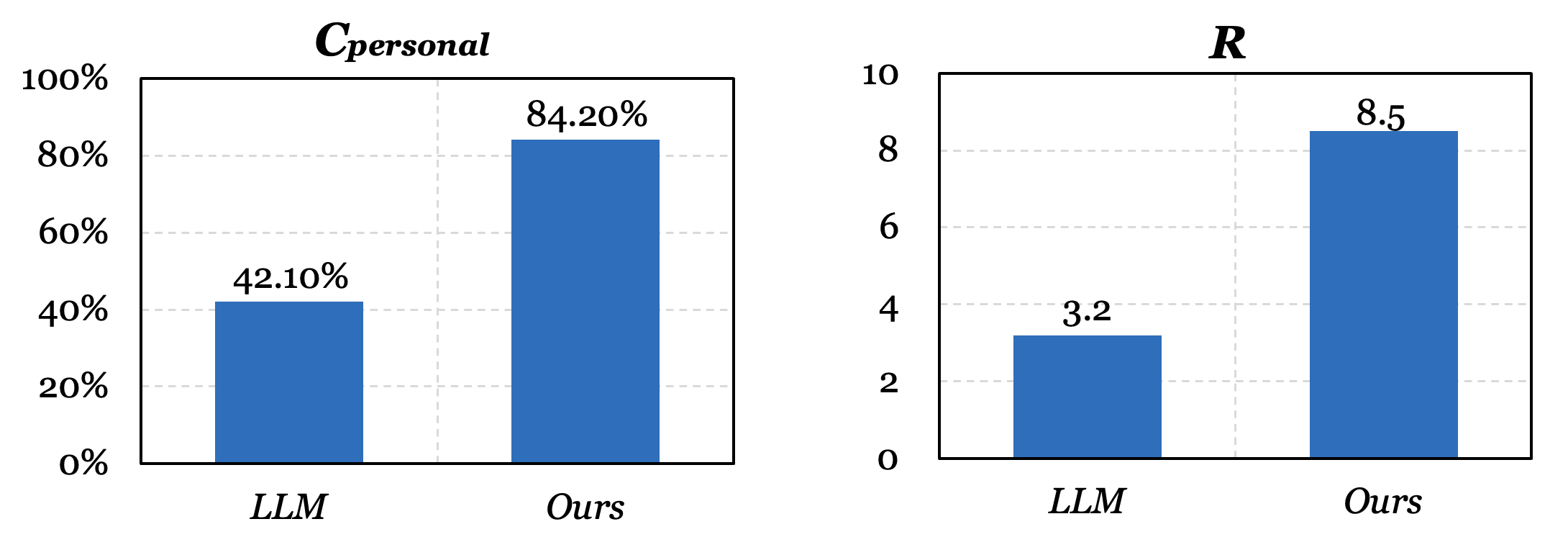} \\ [-4pt]
	\hline
	\addstackgap[7pt]{{\usefont{T1}{ptm}{b}{n}Figure \ref{fig:res_personal}.\hspace{0.09cm}Comparison between LLM and Our Method for Personal Questionnaires}} \\
	\hline
	\end{array}
	\]
	\captionsetup{labelformat=empty}
	\caption{}\label{fig:res_personal}
\end{figure}

For a quantitative comparison of personal questionnaire generation performance, we conduct experiments on the collected 300 EMRs. 
Generated questionnaires are evaluated by medical experts based on the defined metrics. 
Figure~\ref{fig:res_personal} presents the comparative results for Personal Key Fact Coverage ($C_{\text{personal}}$) and Diagnostic Relevance ($R$). 
Our proposed framework achieves a significantly higher mean $C_{\text{personal}}$ (84.2\%) compared to the direct LLM approach (42.10\%). 
Similarly, our method demonstrates markedly superior Diagnostic Relevance ($R$), scoring 8.5 versus 3.2 for the direct LLM. 
The consistently strong performance of our framework across both factual coverage and clinical relevance metrics provides quantitative evidence of the significant advantage of our structured, multi-stage approach in effectively capturing EMR facts and generating clinically pertinent questions, especially when compared to direct LLM prompting.

\textbf{Performance Evaluation for Disease-Specific Questionnaires (RQ3)}

\begin{figure}[htbp]
	\[
	\begin{array}{|c|}
	\hline \\ [-11pt]
	\includegraphics[scale = 0.3]{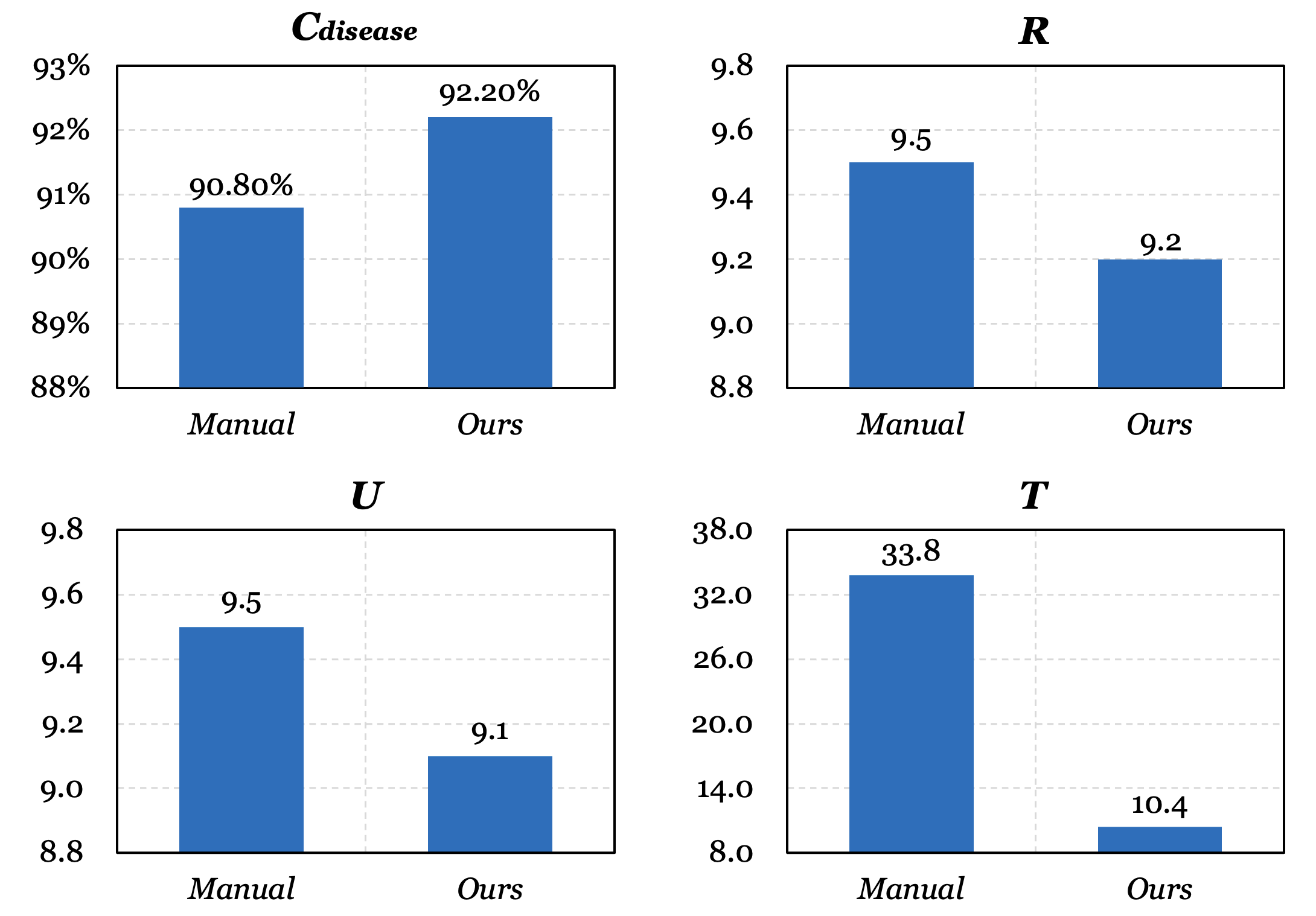} \\ [-4pt]
	\hline
	\addstackgap[7pt]{{\usefont{T1}{ptm}{b}{n}Figure \ref{fig:res_disease}.\hspace{0.09cm}Comparison between Manual and Our Method for Disease Questionnaires}} \\
	\hline
	\end{array}
	\]
	\captionsetup{labelformat=empty}
	\caption{}\label{fig:res_disease}
\end{figure}

We also evaluate the performance of our framework in generating disease-specific pre-consultation questionnaires, comparing it against manual generation by clinical experts. 
Only the manually generated method is used as the baseline because each disease may contain a large number of EMRs, thus exceeding the input token length limit of LLM.
Figure~\ref{fig:res_disease} presents the comparative results for key metrics. 
Our method achieves slightly higher disease-specific key fact coverage ($C_{\text{disease}} = 92.20\%$) compared to manual generation ($90.80\%$). 
While our method demonstrates slightly lower scores for diagnostic relevance ($R = 9.2$) and understandability ($U = 9.1$) compared to the manual baseline ($R = 9.5$, $U = 9.5$), it is important to note that both scores remain high (above 9 on a 10-point scale).
Crucially, compared to the generation time $T$ of the manual process ($33.8$ min), our automated framework drastically reduces it to 10.4 min. 
These results highlight that our proposed automated framework effectively balances achieving high coverage and quality with providing significant improvements in efficiency, presenting a viable and automated alternative to traditional manual questionnaire generation for disease-specific contexts.

\vspace{-0.2cm}
\vspace{-0.9cm}

\vspace{-0.2cm}
\vspace{-0.9cm} 

\section{Conclusion and Discussion}
\label{sec:conclusion}

In this paper, we address the challenge of generating comprehensive pre-consultation questionnaires from complex and voluminous EMRs using LLMs by proposing a novel multi-stage framework leveraging structured knowledge. 
Key methodological contributions include extracting atomic assertions, constructing personal causal networks, and synthesizing disease-specific knowledge by clustering these networks.
The framework automates personalized and disease-specific questionnaire generation, achieving superior performance over baselines in coverage, relevance, understandability, and efficiency, demonstrating significant practical potential for enhancing patient information collection. 
Identified limitations involve dependence on LLM capabilities and the quality and representativeness of training data. 
Future work aims to improve extraction robustness and explore alternative methodologies. 
A crucial next step involves acquiring and utilizing a larger and more diverse dataset of EMRs to enable robust clinical validation of the framework in real-world settings and explore its broader applicability.

\printbibliography

@inproceedings{li2024beyond,
  title={Beyond the Waiting Room: Patient's Perspectives on the Conversational Nuances of Pre-Consultation Chatbots},
  author={Li, Brenna and Gross, Ofek and Crampton, Noah and Kapoor, Mamta and Tauseef, Saba and Jain, Mohit and Truong, Khai N and Mariakakis, Alex},
  booktitle={Proceedings of the 2024 CHI Conference on Human Factors in Computing Systems},
  pages={1--24},
  year={2024}
}

@inproceedings{
    winston2024medical,
    title={Medical question-generation for pre-consultation with {LLM} in-context learning},
    author={Caleb Winston and Cleah Winston and Claris Winston and Chloe Winston},
    booktitle={GenAI for Health: Potential, Trust and Policy Compliance},
    year={2024},
    %url={https://openreview.net/forum?id=du26Irf5kE}
}

@article{guyatt1992evidence,
  title={Evidence-based medicine: a new approach to teaching the practice of medicine},
  author={Guyatt, Gordon and Cairns, John and Churchill, David and Cook, Deborah and Haynes, Brian and Hirsh, Jack and Irvine, Jan and Levine, Mark and Levine, Mitchell and Nishikawa, Jim and others},
  journal={jama},
  volume={268},
  number={17},
  pages={2420--2425},
  year={1992},
  publisher={American Medical Association}
}

@book{world2023tracking,
  title={Tracking universal health coverage: 2023 global monitoring report},
  author={WHO},
  year={2023},
  publisher={World Health Organization},
  url={https://iris.who.int/bitstream/handle/10665/374059/9789240080379-eng.pdf}
}

@book{coallier2017new,
 title={New Patient Intake Form Logbook: Medical Bookkeeping Forms Book},
 author={Coallier, Julien},
 year={2017},
 publisher={CreateSpace Independent Publishing Platform}
}

@inproceedings{ahsan2022machine,
  title={Machine-learning-based disease diagnosis: A comprehensive review},
  author={Ahsan, Md Manjurul and Luna, Shahana Akter and Siddique, Zahed},
  booktitle={Healthcare},
  volume={10},
  number={3},
  pages={541},
  year={2022},
  organization={MDPI}
}

@article{li2024unioqa,
  title={Unioqa: A unified framework for knowledge graph question answering with large language models},
  author={Li, Zhuoyang and Deng, Liran and Liu, Hui and Liu, Qiaoqiao and Du, Junzhao},
  journal={arXiv preprint arXiv:2406.02110},
  year={2024}
}

@inproceedings{park2021knowledge,
  title={Knowledge graph-based question answering with electronic health records},
  author={Park, Junwoo and Cho, Youngwoo and Lee, Haneol and Choo, Jaegul and Choi, Edward},
  booktitle={Machine Learning for Healthcare Conference},
  pages={36--53},
  year={2021},
  organization={PMLR}
}

@article{qiu2024llm,
  title={LLM-based agentic systems in medicine and healthcare},
  author={Qiu, Jianing and Lam, Kyle and Li, Guohao and Acharya, Amish and Wong, Tien Yin and Darzi, Ara and Yuan, Wu and Topol, Eric J},
  journal={Nature Machine Intelligence},
  volume={6},
  number={12},
  pages={1418--1420},
  year={2024},
  publisher={Nature Publishing Group}
}

@inproceedings{dong2024survey,
  title={A Survey on In-context Learning},
  author={Dong, Qingxiu and Li, Lei and Dai, Damai and Zheng, Ce and Ma, Jingyuan and Li, Rui and Xia, Heming and Xu, Jingjing and Wu, Zhiyong and Chang, Baobao and others},
  booktitle={Proceedings of the 2024 Conference on Empirical Methods in Natural Language Processing},
  pages={1107--1128},
  year={2024}
}

@article{who1992international,
  title={International classification of diseases},
  author={WHO and OMS},
  journal={WHO [Internet]},
  year={1992}
}

@misc{youdao_bcembedding_2023,
    title={BCEmbedding: Bilingual and Crosslingual Embedding for RAG},
    author={NetEase Youdao, Inc.},
    year={2023},
    howpublished={\url{https://github.com/netease-youdao/BCEmbedding}},
    url={https://github.com/netease-youdao/BCEmbedding}
}

@inproceedings{yaddaden2023machine,
  title={Machine Learning-Based Pre-Diagnosis Tools in Emergency Departments: Predicting Hospitalization, Mortality and Triage Acuity},
  author={Yaddaden, Yacine and Benahmed, Yacine and Rioux, Marc-Denis and Kallel, Mariem},
  booktitle={2023 IEEE Third International Conference on Signal, Control and Communication (SCC)},
  pages={1--6},
  year={2023},
  organization={IEEE}
}

@inproceedings{likhitha2023developing,
  title={Developing a Pre-Consultation System using Machine Learning for Medical Diagnostics},
  author={Likhitha, Medisetty and Kalyani, G and Vennela, Tulluri Naga and Paul, Dharmapuri Mahith},
  booktitle={2023 7th International Conference on Computing Methodologies and Communication (ICCMC)},
  pages={257--262},
  year={2023},
  organization={IEEE}
}

@article{yuan2022doctor,
  title={Doctor recommendation on healthcare consultation platforms: an integrated framework of knowledge graph and deep learning},
  author={Yuan, Hui and Deng, Weiwei},
  journal={Internet Research},
  volume={32},
  number={2},
  pages={454--476},
  year={2022},
  publisher={Emerald Publishing Limited}
}

@article{abdul2024improving,
  title={Improving preliminary clinical diagnosis accuracy through knowledge filtering techniques in consultation dialogues},
  author={Abdul, Ashu and Chen, Binghong and Phani, Siginamsetty and Chen, Jenhui},
  journal={Computer Methods and Programs in Biomedicine},
  volume={246},
  pages={108051},
  year={2024},
  publisher={Elsevier}
}

@article{varshney2023knowledge,
  title={Knowledge graph assisted end-to-end medical dialog generation},
  author={Varshney, Deeksha and Zafar, Aizan and Behera, Niranshu Kumar and Ekbal, Asif},
  journal={Artificial Intelligence in Medicine},
  volume={139},
  pages={102535},
  year={2023},
  publisher={Elsevier}
}

@article{rotmensch2017learning,
  title={Learning a health knowledge graph from electronic medical records},
  author={Rotmensch, Maya and Halpern, Yoni and Tlimat, Abdulhakim and Horng, Steven and Sontag, David},
  journal={Scientific reports},
  volume={7},
  number={1},
  pages={5994},
  year={2017},
  publisher={Nature Publishing Group UK London}
}

@article{bean2017knowledge,
  title={Knowledge graph prediction of unknown adverse drug reactions and validation in electronic health records},
  author={Bean, Daniel M and Wu, Honghan and Iqbal, Ehtesham and Dzahini, Olubanke and Ibrahim, Zina M and Broadbent, Matthew and Stewart, Robert and Dobson, Richard JB},
  journal={Scientific reports},
  volume={7},
  number={1},
  pages={16416},
  year={2017},
  publisher={Nature Publishing Group UK London}
}

@article{shang2024electronic,
  title={Electronic health record--oriented knowledge graph system for collaborative clinical decision support using multicenter fragmented medical data: design and application study},
  author={Shang, Yong and Tian, Yu and Lyu, Kewei and Zhou, Tianshu and Zhang, Ping and Chen, Jianghua and Li, Jingsong},
  journal={Journal of Medical Internet Research},
  volume={26},
  pages={e54263},
  year={2024},
  publisher={JMIR Publications Toronto, Canada}
}

@article{gazzotti2022extending,
  title={Extending electronic medical records vector models with knowledge graphs to improve hospitalization prediction},
  author={Gazzotti, Rapha{\"e}l and Faron, Catherine and Gandon, Fabien and Lacroix-Hugues, Virginie and Darmon, David},
  journal={Journal of Biomedical Semantics},
  volume={13},
  number={1},
  pages={6},
  year={2022},
  publisher={Springer}
}

@article{thirunavukarasu2023large,
  title={Large language models in medicine},
  author={Thirunavukarasu, Arun James and Ting, Darren Shu Jeng and Elangovan, Kabilan and Gutierrez, Laura and Tan, Ting Fang and Ting, Daniel Shu Wei},
  journal={Nature medicine},
  volume={29},
  number={8},
  pages={1930--1940},
  year={2023},
  publisher={Nature Publishing Group US New York}
}

@article{clusmann2023future,
  title={The future landscape of large language models in medicine},
  author={Clusmann, Jan and Kolbinger, Fiona R and Muti, Hannah Sophie and Carrero, Zunamys I and Eckardt, Jan-Niklas and Laleh, Narmin Ghaffari and L{\"o}ffler, Chiara Maria Lavinia and Schwarzkopf, Sophie-Caroline and Unger, Michaela and Veldhuizen, Gregory P and others},
  journal={Communications medicine},
  volume={3},
  number={1},
  pages={141},
  year={2023},
  publisher={Nature Publishing Group UK London}
}

@article{lievin2024can,
  title={Can large language models reason about medical questions?},
  author={Li{\'e}vin, Valentin and Hother, Christoffer Egeberg and Motzfeldt, Andreas Geert and Winther, Ole},
  journal={Patterns},
  volume={5},
  number={3},
  year={2024},
  publisher={Elsevier}
}

@article{lee2020biobert,
  title={BioBERT: a pre-trained biomedical language representation model for biomedical text mining},
  author={Lee, Jinhyuk and Yoon, Wonjin and Kim, Sungdong and Kim, Donghyeon and Kim, Sunkyu and So, Chan Ho and Kang, Jaewoo},
  journal={Bioinformatics},
  volume={36},
  number={4},
  pages={1234--1240},
  year={2020},
  publisher={Oxford University Press}
}

@article{huang2019clinicalbert,
  title={Clinicalbert: Modeling clinical notes and predicting hospital readmission},
  author={Huang, Kexin and Altosaar, Jaan and Ranganath, Rajesh},
  journal={arXiv preprint arXiv:1904.05342},
  year={2019}
}

@article{yang2022gatortron,
  title={Gatortron: A large clinical language model to unlock patient information from unstructured electronic health records},
  author={Yang, Xi and Chen, Aokun and PourNejatian, Nima and Shin, Hoo Chang and Smith, Kaleb E and Parisien, Christopher and Compas, Colin and Martin, Cheryl and Flores, Mona G and Zhang, Ying and others},
  journal={arXiv preprint arXiv:2203.03540},
  year={2022}
}

@article{mcduff2025towards,
  title={Towards accurate differential diagnosis with large language models},
  author={McDuff, Daniel and Schaekermann, Mike and Tu, Tao and Palepu, Anil and Wang, Amy and Garrison, Jake and Singhal, Karan and Sharma, Yash and Azizi, Shekoofeh and Kulkarni, Kavita and others},
  journal={Nature},
  pages={1--7},
  year={2025},
  publisher={Nature Publishing Group UK London}
}

@article{tripathi2024efficient,
  title={Efficient healthcare with large language models: optimizing clinical workflow and enhancing patient care},
  author={Tripathi, Satvik and Sukumaran, Rithvik and Cook, Tessa S},
  journal={Journal of the American Medical Informatics Association},
  volume={31},
  number={6},
  pages={1436--1440},
  year={2024},
  publisher={Oxford University Press}
}

@article{tang2023evaluating,
  title={Evaluating large language models on medical evidence summarization},
  author={Tang, Liyan and Sun, Zhaoyi and Idnay, Betina and Nestor, Jordan G and Soroush, Ali and Elias, Pierre A and Xu, Ziyang and Ding, Ying and Durrett, Greg and Rousseau, Justin F and others},
  journal={NPJ digital medicine},
  volume={6},
  number={1},
  pages={158},
  year={2023},
  publisher={Nature Publishing Group UK London}
}

@incollection{bran2024transformers,
  title={Transformers and large language models for chemistry and drug discovery},
  author={Bran, Andres M and Schwaller, Philippe},
  booktitle={Drug Development Supported by Informatics},
  pages={143--163},
  year={2024},
  publisher={Springer}
}

@book{zhang2002association,
  title={Association rule mining: models and algorithms},
  author={Zhang, Chengqi and Zhang, Shichao},
  year={2002},
  publisher={Springer}
}

@article{urbanowicz2009learning,
  title={Learning classifier systems: a complete introduction, review, and roadmap},
  author={Urbanowicz, Ryan J and Moore, Jason H},
  journal={Journal of Artificial Evolution and Applications},
  volume={2009},
  number={1},
  pages={736398},
  year={2009},
  publisher={Wiley Online Library}
}

@book{de2002artificial,
  title={Artificial immune systems: a new computational intelligence approach},
  author={De Castro, Leandro Nunes and Timmis, Jonathan},
  year={2002},
  publisher={Springer Science \& Business Media}
}

@article{ni2024knowledge,
  title={Knowledge graph and deep learning-based text-to-GraphQL model for intelligent medical consultation chatbot},
  author={Ni, Pin and Okhrati, Ramin and Guan, Steven and Chang, Victor},
  journal={Information Systems Frontiers},
  volume={26},
  number={1},
  pages={137--156},
  year={2024},
  publisher={Springer}
}

@article{tai2017electronic,
  title={Electronic health record logs indicate that physicians split time evenly between seeing patients and desktop medicine},
  author={Tai-Seale, Ming and Olson, Cliff W and Li, Jinnan and Chan, Albert S and Morikawa, Criss and Durbin, Meg and Wang, Wei and Luft, Harold S},
  journal={Health affairs},
  volume={36},
  number={4},
  pages={655--662},
  year={2017}
}

@article{johnson2021precision,
  title={Precision medicine, AI, and the future of personalized health care},
  author={Johnson, Kevin B and Wei, Wei-Qi and Weeraratne, Dilhan and Frisse, Mark E and Misulis, Karl and Rhee, Kyu and Zhao, Juan and Snowdon, Jane L},
  journal={Clinical and translational science},
  volume={14},
  number={1},
  pages={86--93},
  year={2021},
  publisher={Wiley Online Library}
}

@article{murtagh2014ward,
  title={Ward’s hierarchical agglomerative clustering method: which algorithms implement Ward’s criterion?},
  author={Murtagh, Fionn and Legendre, Pierre},
  journal={Journal of classification},
  volume={31},
  pages={274--295},
  year={2014},
  publisher={Springer}
}

@misc{OpenAI_GPT4o,
  author = {{OpenAI}}, 
  title = {GPT-4o}, 
  url = {https://openai.com/index/hello-gpt-4o/},
  year = {2024}
}

\end{document}